\documentclass[letterpaper,10pt,conference]{ieeeconf}

\IEEEoverridecommandlockouts
\usepackage[hidelinks]{hyperref}
\usepackage{graphicx}
\usepackage{amsmath}
\usepackage{amssymb}
\usepackage{algorithm,algorithmicx}
\usepackage{algpseudocode}
\usepackage{cleveref}
\usepackage{xspace}
\usepackage{xcolor}
\usepackage{subcaption}
\usepackage{cite}
\usepackage{tablefootnote}
\usepackage{makecell}
\usepackage{flushend}
\newcommand{\MS}{Motion~Stack\xspace}
\newcommand{\MB}{MoonBot\xspace}
\newcommand{\MBs}{MoonBots\xspace}

\newcommand{\moit}{\textit{Moveit!}\xspace}
\newcommand{\rt}{ROS2\xspace}

\crefname{section}{Sec.}{Secs.}
\crefname{figure}{Fig.}{Figs.}
\crefname{equation}{}{}
\crefname{algorithm}{Alg.}{Algs.}
\crefname{table}{Tab.}{Tabs.}

\title{\LARGE \bf
An Agnostic End-Effector Alignment Controller \\
for Robust Assembly of Modular Space Robots
}

\author{Shamistan Karimov$^{1}$, Elian Neppel$^{1}$,
Shreya Santra$^{1}$, Kentaro Uno$^{1}$, and Kazuya Yoshida$^{1}$% <-this % stops a space
\thanks{
*This work was supported by JST SPRING, Grant Number JPMJSP2114 and JST Moonshot R\&D Program, Grant Number JPMJMS223B.
    }%
\thanks{$^{1}$S. Karimov, E. Neppel, S. Santra, K. Uno, and K. Yoshida are with the Space Robotics Lab (SRL), Dept.\ of Aerospace Engineering, Tohoku University, Sendai 980–8579, Japan. (E-mail: \tt{karimov.shamistan.p8@dc.tohoku.ac.jp})}
}

\begin{document}
\bstctlcite{BSTcontrol}  % must appear before the first \cite

\maketitle
\thispagestyle{empty}
\pagestyle{empty}

%%%%%%%%%%%%%%%%%%%%%%%%%%%%%%%%%%%%%%%%%%%%%%%%%%%%%%%%%%%%%%%%%%%%%%%%%%%%%%%%
\begin{abstract}
Modular robots offer reconfigurability and fault tolerance essential for lunar missions,
but require controllers that adapt safely to real-world disturbances. We build on our
previous hardware-agnostic actuator synchronization in Motion Stack to develop a new
controller enforcing adaptive velocity bounds via a dynamic hypersphere clamp. Using only real-time end-effector and target pose measurements, the controller adjusts
its translational and rotational speed limits to ensure smooth, stable alignment without
abrupt motions. We implemented two variants -- a discrete, step-based version and a
continuous, velocity-based version -- and tested them on two \MB limbs in JAXA’s
lunar environment simulator. Field trials demonstrate that the step-based variant produces highly predictable, low-wobble
motions, while the continuous variant converges more quickly and maintains millimeter-level
positional accuracy, and both remain robust across limbs with differing mechanical imperfections and sensing noise (e.g., backlash and flex). These results highlight the flexibility and robustness of our
robot-agnostic framework for autonomous self-assembly and reconfiguration under harsh conditions.
\end{abstract}

%%%%%%%%%%%%%%%%%%%%%%%%%%%%%%%%%%%%%%%%%%%%%%%%%%%%%%%%%%%%%%%%%%%%%%%%%%%%%%%%

\section{Introduction}
Space missions demand robotic systems that can withstand extreme conditions,
recover from failures, and perform varied manipulation tasks with minimal
human oversight.
Modular self-assembly enables collections of robotic modules -- each with onboard actuation,
sensing, and docking interfaces -- to autonomously connect and reconfigure into larger structures
tailored to specific tasks or environments \cite{wei_sambot_2011, tan_sambotii_2018}.
Such systems can form “organism-like” assemblies during motion and even perform self-repair by
replacing faulty modules without halting group activities, thereby enhancing reliability in unstructured
or harsh settings \cite{peck_self-assembly_2022}.
By leveraging standard mechanical and electrical interfaces, modular robots can adapt their morphology
on demand -- growing, shrinking, or reshaping to accomplish manipulation, locomotion, or load-bearing
operations \cite{doi:10.1177/1687814016659597, ModularReconfig2020}.
% -------------------------------------------------------------------------------
\begin{figure}[t]
\vspace{2mm}
  \centering
  \includegraphics[width=\columnwidth, trim=0.7cm 0cm 5.5cm 0cm, clip]{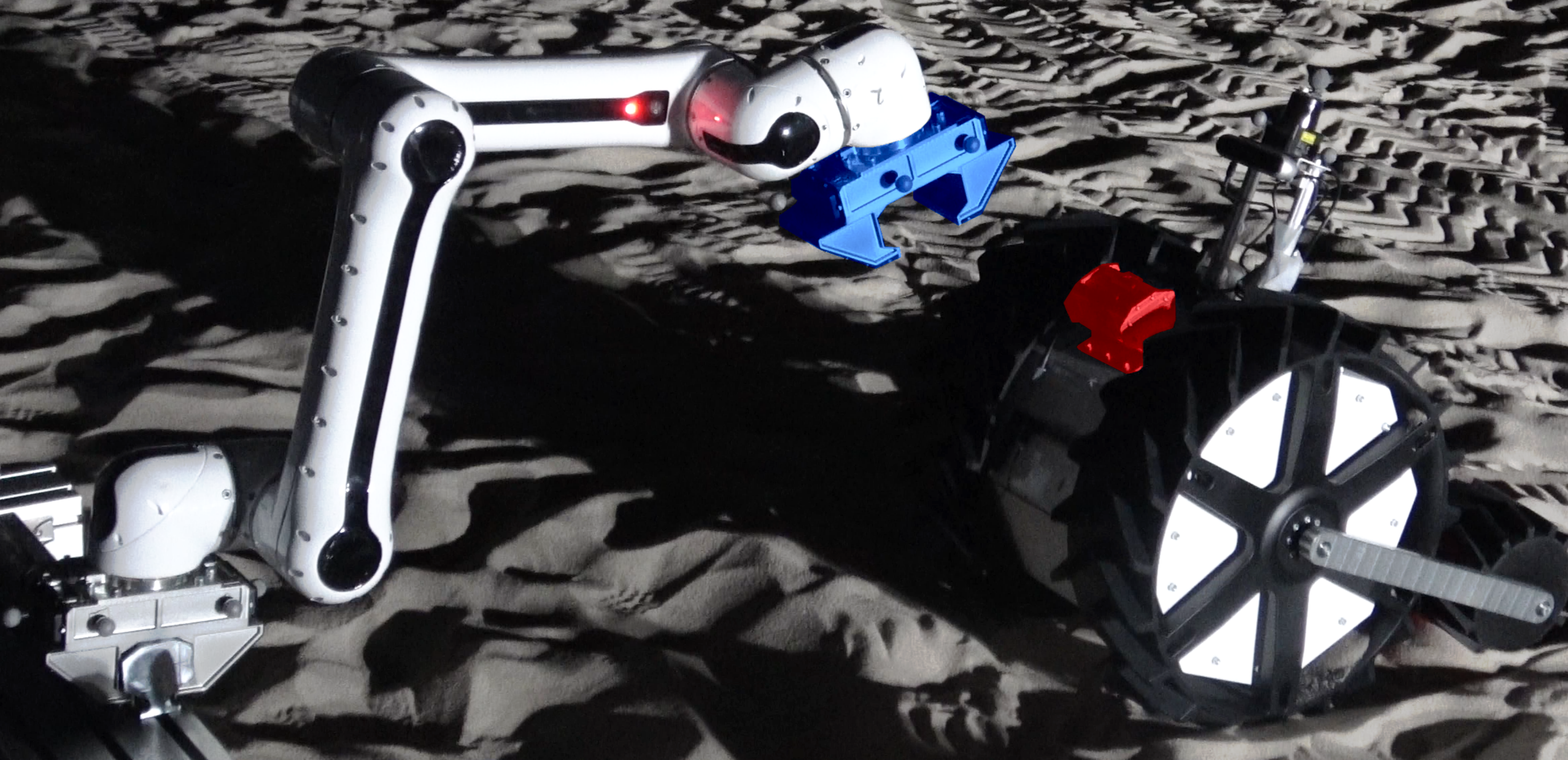}
    \caption{End-effector of the \MB Limb V2 (blue) approaching the grasping point on the \MB Wheel V2 (red).}
  \label{fig:limb2_wheel}
%\vspace{-2mm}
\end{figure}
% -------------------------------------------------------------------------------
As part of the Moonshot R\&D Project (Goal 3) \cite{moonshotgoal3}, we develop
modular robot systems for lunar exploration and outpost construction.
This paper addresses one of the needs --  an autonomous, robust alignment controller that maintains
precise, safe manipulation in the face of unpredictable external disturbances.
The controller adapts its motion commands in real time, ensuring reliable
performance across heterogeneous modular configurations.

Early modular systems (e.g., chain-style and genderless connectors)
showed shape-shifting locomotion and basic manipulation, while more
recent platforms automate model updates and docking
\cite{yim_modular_2002, yim_self-reconfigurable_2003, ModMan2020}.
Space-oriented concepts exist, but in-situ validation and unified
control remain limited \cite{space_reconf2021, moobot_zero, Yim2003ModularRR}.

Existing software for modular robots often assumes idealized conditions and rigid kinematics,
but real-world deployments on lunar‐like terrain reveal significant discrepancies.
Mechanical issues such as joint backlash, structural flex, and connector wobble introduce unpredictable pose errors.
Sensor noise and latency -- exacerbated by dust, temperature extremes,
and dynamic loads -- further degrade open‐loop performance.
Under these harsh conditions, controllers lacking real‐time adaptation risk overshoot, oscillation, or failure to converge.
Surveys of collaborative manipulation frameworks outline high-level
coordination strategies \cite{colab_overview}, yet stop short of offering a
robot-agnostic middleware. Similarly, \rt-centric solutions like the \moit Task
Constructor \cite{MoveIt2019} and the \textit{ros2\_control} framework
\cite{ros2_control} excel at planning and device-agnostic actuation for single
manipulators, but are not designed for multiple modular units performing tasks
like self-assembly or reconfiguration. We were motivated by Storiale et al.’s
ROS-based, hardware-agnostic interaction controllers \cite{robotagnosticros},
which adapt to different robots without reprogramming.

In our recent work \cite{neppel_robust_2025}, we introduced a hypersphere-based
synchronization method within \MS \cite{motionstack} that tightly couples
the commanded multidimensional velocity to the manipulator’s proprioceptive state,
effectively closing the loop on internal joint feedback for trajectory compliance.
However, that implementation did not leverage external pose measurements --
such as vision-based tracking -- to correct for unmodeled flexibility, sensor drift, or backlash.
In contrast, the controller presented here closes the loop around exteroceptive
feedback, using real-time end-effector and target poses to dynamically adjust
its velocity limits and maintain precise, stable alignment.

Our work tackles this gap by embedding external sensor-driven stability metrics and adaptive clamping into the alignment loop,
ensuring robust convergence despite mechanical and environmental variability.
The allowable velocity region adjusts in real time based on pose error and external metrics (e.g.,
jitter), yielding smooth, stable alignment for self-assembly and
reconfiguration tasks. Two controller versions were iteratively developed and
field-tested on two 7-DOF modular \MB limbs \cite{moonbot} (see \cref{fig:limb2_wheel}) at
JAXA’s Advanced Facility for Space Exploration on the Sagamihara campus
\cite{jaxafield}, where a lunar environment-simulated field allowed repeated
trials over three months. These experiments validate the controller’s
precision, responsiveness, and robustness under harsh conditions.

Our implementation builds on \MS, a \rt-based middleware framework for
motion planning and coordination. By developing a new, robot-agnostic
controller layer atop \MS, we achieve seamless integration of the adaptive hypersphere clamp across different manipulator configurations. In the following sections, we detail our controller design, and present the field-test methodology that guided our
iterative refinements.

%%%%%%%%%%%%%%%%%%%%%%%%%%%%%%%%%%%%%%%%%%%%%%%%%%%%%%%%%%%%%%%%%%%%%%%%%%%%%%%%
\section{Method}

\label{sec:method}

The alignment controller operates purely on two inputs provided by external sensors:
the measured pose of the end-effector \((x_e, q_e)\) and the measured pose of the target \((x_t, q_t)\).
From these, it must:

\begin{itemize}
  \item Compute translational and rotational errors robustly, even in the presence of sensor noise.
  \item Drive the manipulator smoothly and safely, avoiding abrupt jumps or oscillations.
  \item Adapt its step size or velocity limits automatically based on current error and stability metrics.
  \item Version 1, wait for each step to complete and settle; for Version 2, ensure the adaptive clamp reduces velocities rather than full stops.
\end{itemize}

At each high-level update, the controller:

\begin{enumerate}
  \item Reads \((x_e,q_e)\) and \((x_t,q_t)\).
  \item Computes the raw errors
  \begin{equation}\label{eq:pose_error}
  \begin{aligned}
    \Delta d &= \|x_t - x_e\|\;,\\
      \Delta\theta &= 2\arccos\bigl(\lvert q_t^{-1} \cdot q_e\rvert\bigr)\;.
  \end{aligned}
  \end{equation}
    where \(x_e,x_t\in\mathbb{R}^3\) are end-effector and target positions, and
    \(q_e,q_t\in\mathbb{S}^3\) are unit quaternions representing current and desired orientations.
  \item Feeds these into one of two alignment strategies.
  \item Relies on a 30 Hz low-level execution loop (Motion Stack) to interpolate or track the commanded poses/velocities.
\end{enumerate}

\vspace{1ex}
\subsection{Controller Version 1: Two-Stage LERP Alignment}
\label{sec:version1}

Controller Version 1, a \emph{naive approach}, issues a sequence of small pose steps
and waits for each to complete and settle before proceeding.

\medskip\noindent\textbf{Error Averaging and Stability Check.}
Before each command, the translational error vector is appended to a short buffer and averaged:
\begin{gather}
  \overline{e}_{xyz} = \frac{1}{N}\sum_{i=1}^{N}(x_t - x_e)_i
  \label{eq:error_avg}
\end{gather}
This reduces the effect of measurement noise. Additionally, we verify that:
\begin{itemize}
  \item The previous step has finished interpolating.
  \item The measured error pose \cref{eq:pose_error} hasn’t changed by more than \(\delta_{\rm jump}\);
    if it has, we hold all motion and re-average using \cref{eq:error_avg}
    until stability returns.
\end{itemize}
Only when both conditions are met do we proceed to the next step.

\medskip\noindent\textbf{Offset Computation.}
While the pose error components in \cref{eq:pose_error} exceed their respective convergence thresholds, we compute:
\begin{gather}
  s =
  \begin{cases}
    s_{\max}, & \Delta d > r_{\mathrm{high}},\\
    s_{\min}, & \Delta d < r_{\mathrm{low}},\\
    s_{\min}
      + \dfrac{\Delta d - r_{\mathrm{low}}}{r_{\mathrm{high}} - r_{\mathrm{low}}}(s_{\max} - s_{\min}),
      & \text{otherwise},
  \end{cases}
  \label{eq:step_size}
\end{gather}
As shown in \cref{eq:step_size}, the step size \(s\) is chosen by linearly interpolating between its minimum and maximum
values based on the current translational error’s position between \(r_{\min}\) and \(r_{\max}\).
Then we form the relative translation offset:
\begin{gather}
  \Delta p = s\,\frac{\overline{e}_{xyz}}{\|\overline{e}_{xyz}\|},
  \label{eq:coarse_delta_p}
\end{gather}
In the equation \cref{eq:coarse_delta_p}, \(\Delta p\) is the averaged error vector \cref{eq:error_avg},
scaled by the step size \cref{eq:step_size}, thus pointing toward the target.
We then clamp the incremental rotation step to lie between \(\theta_{\min}\) and \(\theta_{\max}\), and send the resulting
incremental pose command to Motion Stack’s Linear Interpolation (LERP)-based inverse kinematics synchronizer,
which ensures smooth, incremental joint‐space motion until convergence.

\vspace{1ex}
\subsection{Controller Version 2: Continuous Adaptive Hypersphere Clamp}
\label{sec:version2}

Controller Version 2 departs from discrete steps and instead issues a continuous 6-DOF velocity command.
At each high-level tick, it adaptively bounds translation and rotation based on the current error and stability.

\medskip\noindent\textbf{Base Radii Interpolation.}
We map \(\Delta d\) and \(\Delta\theta\) (from \cref{eq:pose_error}) to translation/rotation
bounds via clamped, piecewise-linear interpolation:
\begin{equation}
\begin{aligned}
\Delta_t &=
\begin{cases}
  t_{\min}, & \Delta d \le D_{\rm near},\\
  t_{\max}, & \Delta d \ge D_{\rm far},\\
  t_{\min} + \alpha\,(t_{\max}-t_{\min}), & \text{otherwise},
\end{cases}\\[2mm]
\Delta_r &=
\begin{cases}
  r_{\min}, & \Delta\theta \le A_{\rm near},\\
  r_{\max}, & \Delta\theta \ge A_{\rm far},\\
  r_{\min} + \beta\,(r_{\max}-r_{\min}), & \text{otherwise},
\end{cases}
\end{aligned}
\label{eq:base_radii}
\end{equation}
with interpolation weights \cref{eq:interp_weights}
\begin{equation}
  \alpha=\frac{\Delta d - D_{\rm near}}{D_{\rm far}-D_{\rm near}},
  \qquad
  \beta =\frac{\Delta\theta - A_{\rm near}}{A_{\rm far}-A_{\rm near}}.
\label{eq:interp_weights}
\end{equation}

In \cref{eq:base_radii}, \(D_{\rm near/far}\) and \(A_{\rm near/far}\) define the input
windows over which \((t_{\min}\!\to\!t_{\max})\) and \((r_{\min}\!\to\!r_{\max})\) are blended;
\cref{eq:interp_weights} gives the linear weights. Inputs below the “near” thresholds
select the minimum bounds, above the “far” thresholds select the maximum, and values
in between are interpolated linearly.

\medskip\noindent\textbf{Shrink Factors.}
Raw radii from \cref{eq:base_radii} do not account for real-world instability.
We therefore compute three shrink factors that penalize jitter (\(f_j\)), large error changes (\(f_k\)),
and off-axis motion (\(f_s\)):

\begin{equation}
\begin{aligned}
  f_j &= \frac{1}{1 + \alpha_j\,\sigma_{\rm jitter}},\\
  f_k &= \frac{1}{1 + \alpha_k\,\lvert \Delta d_k - \Delta d_{k-1}\rvert},\\
  f_s &= \max\!\bigl(\epsilon,\;1 - \tfrac{\mathrm{RMS}_\perp}{\tau_{\rm jitter}}\bigr),
\end{aligned}
\label{eq:shrink_factors}
\end{equation}

Each shrink factor in \cref{eq:shrink_factors} is designed to lie in \((0,1]\) and is governed
by a small set of tunable constants:

\begin{itemize}
  \item \(\alpha_j\) weights the impact of measured sensor jitter \(\sigma_{\rm jitter}\).
    A larger \(\alpha_j\) makes \(f_j\) decrease more rapidly as noise increases,
    shrinking the velocity envelope under high-noise conditions.
  \item \(\alpha_k\) sets how strongly we react to \emph{rapid changes in distance error}
      between ticks, \(|\Delta d_k-\Delta d_{k-1}|\).
      Larger \(\alpha_k\) makes the controller slow down more when the error jumps
      (e.g., overshoot or reversals), while steady, gradual changes keep \(f_k \approx 1\).
  \item \(\mathrm{RMS}_\perp\) is the root-mean-square of the end-effector’s motion perpendicular to
    the main error direction, computed over a short history window.
  \item \(\tau_{\rm jitter}\) sets the characteristic threshold for off-axis motion:
    when \(\mathrm{RMS}_\perp\) approaches \(\tau_{\rm jitter}\), \(f_s\) falls toward its lower bound.
  \item \(\epsilon\) is a small constant (e.g., 0.01) that prevents \(f_s\) from collapsing to zero,
    ensuring the controller never fully freezes its velocity.
\end{itemize}

By carefully selecting \(\alpha_j,\alpha_k,\tau_{\rm jitter},\) and \(\epsilon\), the controller can be
tuned to match the noise characteristics and dynamic behavior of different robot limbs and sensing setups.

\medskip\noindent\textbf{Effective Radii.}
We combine base radii and shrink factors to obtain the effective bounds:

\begin{equation}
  \Delta_t^{\rm eff} = \Delta_t \,f_j\,f_k\,f_s,
  \quad
  \Delta_r^{\rm eff} = \Delta_r \,f_j\,f_k.
\label{eq:eff_radii}
\end{equation}

By construction \cref{eq:eff_radii}, these effective radii shrink whenever any instability
metric grows, ensuring conservative commands under noisy or dynamic conditions.

\medskip\noindent\textbf{Raw Velocity Computation.}
Using the latest end-effector and target poses, we form the “raw” 6-D velocity:

\begin{equation}
  v_{\rm raw} = x_t - x_e,
  \qquad
  \omega_{\rm raw} = \mathrm{axis}\bigl(q_e^{-1}q_t\bigr)\,\Delta\theta.
\label{eq:raw_velocity}
\end{equation}

From \cref{eq:raw_velocity}, \(v_{\rm raw}\) is the translation vector toward the target, and \(\omega_{\rm raw}\)
is the rotation vector aligned with the quaternion error axis.

\medskip\noindent\textbf{Ellipsoidal Clamping.}
We normalize the raw 6-vector \cref{eq:raw_velocity} against the effective radii to enforce the hypersphere constraint:

\begin{equation}
  u =
  \begin{bmatrix}
    v_{\rm raw}/\Delta_t^{\rm eff}\\[3pt]
    \omega_{\rm raw}/\Delta_r^{\rm eff}
  \end{bmatrix},
  \quad
  \hat u = \frac{u}{\max\bigl(1,\|u\|\bigr)},
  \label{eq:ellipsoid_norm}
\end{equation}
then recover the clamped velocities:
\begin{equation}
  v_c = \hat u_{1:3}\,\Delta_t^{\rm eff},
  \quad
  \omega_c = \hat u_{4:6}\,\Delta_r^{\rm eff}.
  \label{eq:clamped_vel}
\end{equation}

By clamping the 6-D velocity vector within a single unit hypersphere (cf.\ \cref{eq:ellipsoid_norm}),
we inherently balance translational and rotational motions: neither component can exceed its
own effective radius disproportionately. As a result, both the positional error \(\Delta d\) and
orientation error \(\Delta\theta\) decay together in a coordinated fashion, yielding smooth,
monotonic convergence of the end-effector pose toward the target without abrupt dominance of
either linear or angular motion.

\medskip\noindent\textbf{Temporal Smoothing.}
To avoid abrupt changes between controller ticks, we low-pass filter the clamped velocities:

\begin{equation}
\begin{aligned}
  v_{k+1} &= v_k + \frac{\Delta t}{\tau_{\rm lin} + \Delta t}\,(v_c - v_k),\\
  \omega_{k+1} &= \omega_k + \frac{\Delta t}{\tau_{\rm rot} + \Delta t}\,(\omega_c - \omega_k).
\end{aligned}
\label{eq:smooth}
\end{equation}

As shown in \cref{eq:smooth}, the filter blends the raw clamped commands \((v_c,\omega_c)\)
with the previous output \((v_k,\omega_k)\), trading off responsiveness (smaller \(\tau\))
against smoothness (larger \(\tau\)).

With these steps, Version 2 continuously adapts to both the magnitude of the pose error and
the measured stability of the system, producing smooth yet assertive alignment commands.

In practice, the choice of the weights in
\cref{eq:shrink_factors} sets a continuum between “Version 1-like” smoothness and
“speed-first” behavior; for the reported results we selected the latter.

% \subsection{Alignment Controller's core loop overview.}
%
% \cref{alg:alignment_final} summarizes the core alignment logic shared by both controllers.
% Controller Version 1 employs a straightforward, whereas Version 2 augments this pipeline
% with adaptive scaling factors and stability metrics to modulate its velocity commands in real time.
%
%
% \begin{algorithm}[ht]
%   \caption{Alignment Logic}
%   \label{alg:alignment_final}
%   \begin{algorithmic}[1]
%     \Procedure{AlignController}{version}
%       \If{version == 1}
%         \While{\(\Delta d > d_{\mathrm{conv}}\) or \(\Delta\theta > \theta_{\mathrm{conv}}\)}
%           \State compute and send LERP offset, wait for completion and settling
%         \EndWhile
%       \Else
%         \While{\(\Delta d > d_{\mathrm{conv}}\) or \(\Delta\theta > \theta_{\mathrm{conv}}\)}
%           \State compute \(\Delta_t^{\rm eff},\,\Delta_r^{\rm eff}\)
%           \State compute and clamp \((v_{\rm raw},\omega_{\rm raw})\to(v_c,\omega_c)\)
%           \State low-pass filter and execute at 30 Hz
%         \EndWhile
%       \EndIf
%       \State stop all motion
%     \EndProcedure
%   \end{algorithmic}
% \end{algorithm}

%%%%%%%%%%%%%%%%%%%%%%%%%%%%%%%%%%%%%%%%%%%%%%%%%%%%%%%%%%%%%%%%%%%%%%%%%%%%%%%%
\section{Experiments and Results}
\label{sec:experiment}

\subsection{Self-Assembly}
\begin{figure}[tb]
  \vspace{2mm}
  \centering
  \begin{subfigure}[b]{0.49\columnwidth}
    \centering
    \includegraphics[width=\linewidth,height=0.6\linewidth, trim={60 0 300 0}, clip]{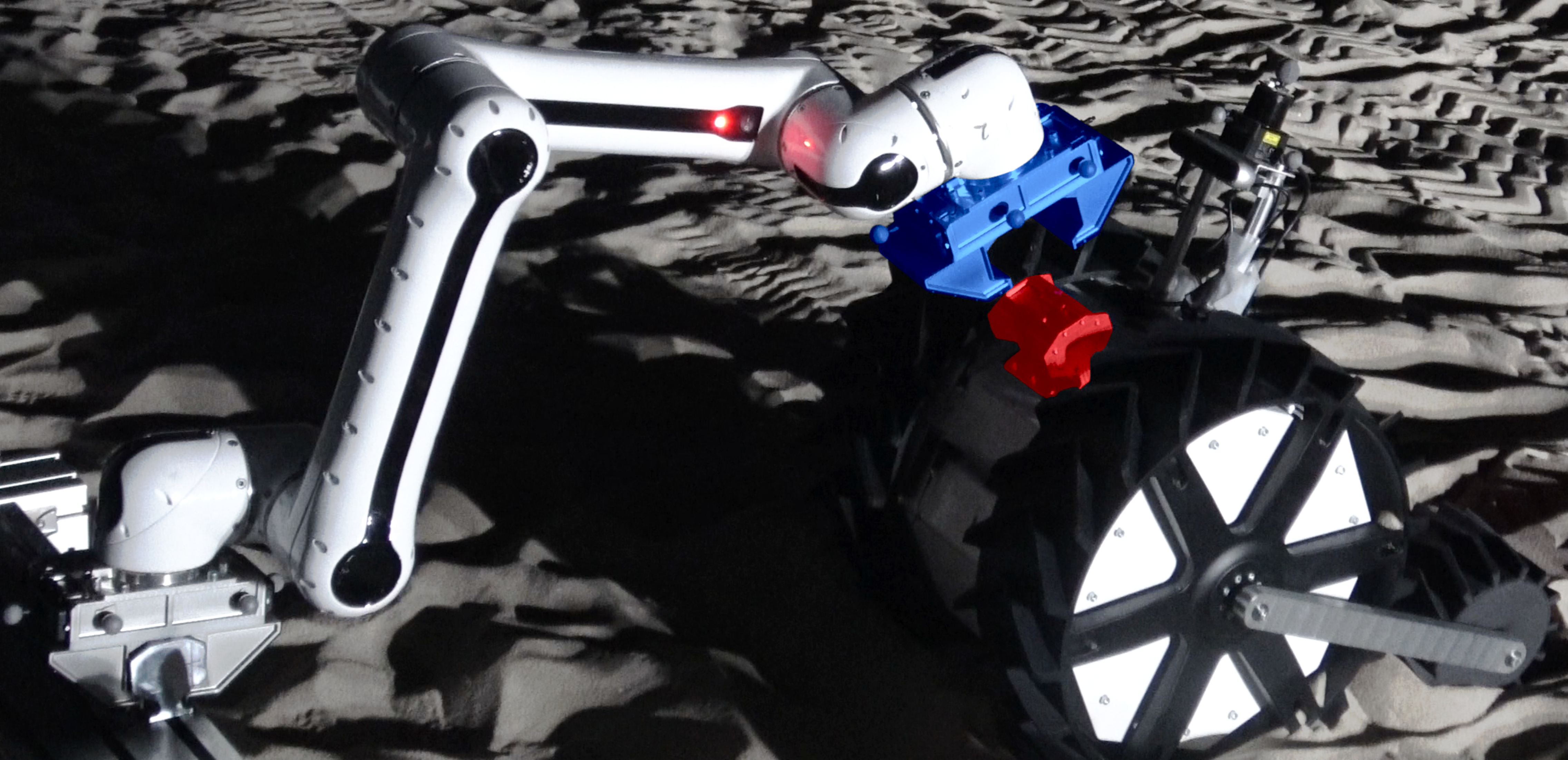}
    \caption{Alignment}
    \label{fig:exp_align}
  \end{subfigure}
  \hfill
  \begin{subfigure}[b]{0.49\columnwidth}
    \centering
    \includegraphics[width=\linewidth, height=0.6\linewidth, trim={60 0 300 0}, clip]{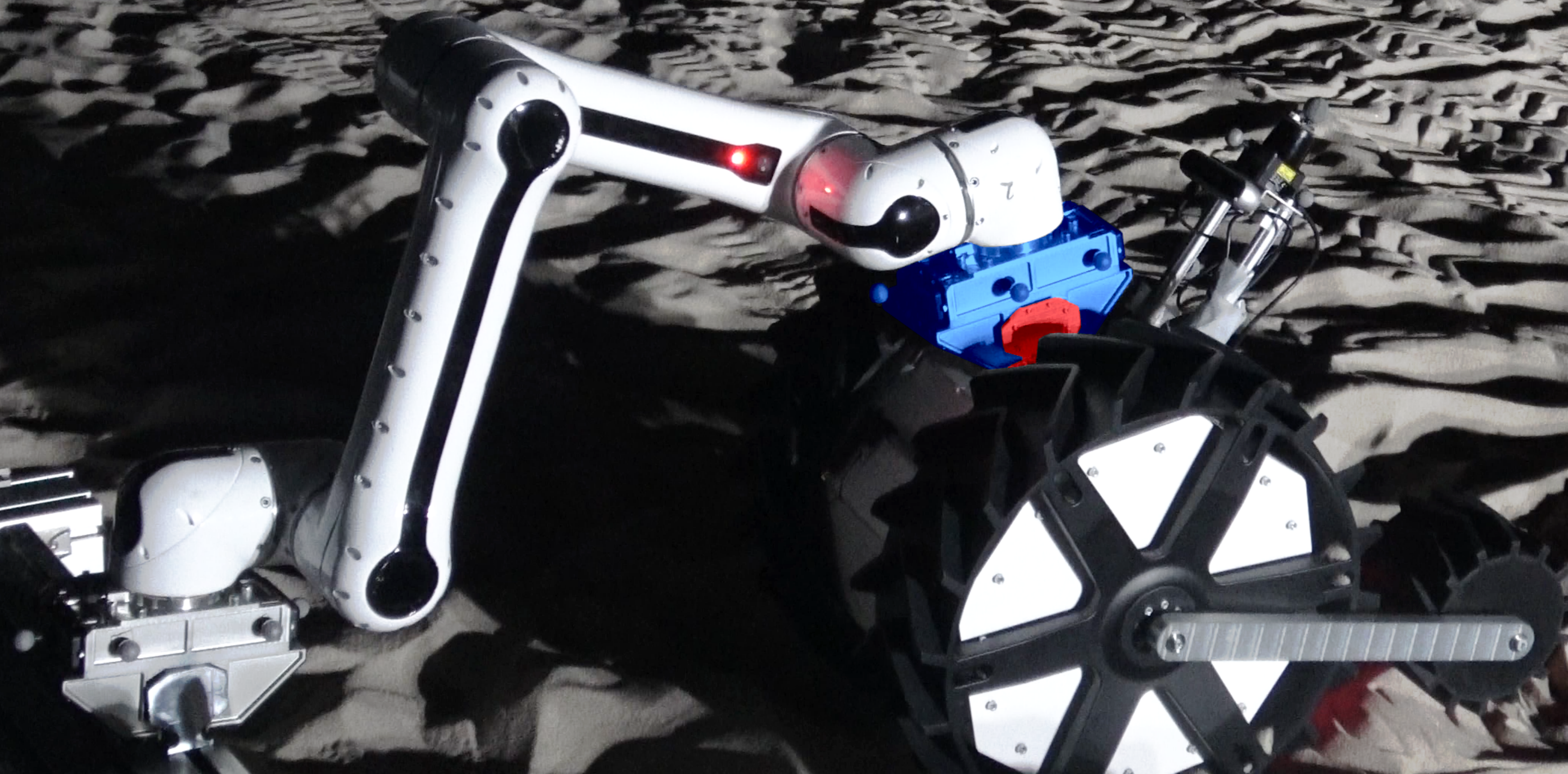}
    \caption{Grasping}
    \label{fig:exp_grasp}
  \end{subfigure}
  \hfill
  % \vspace{.5mm}
  \begin{subfigure}[b]{1.0\columnwidth}
    \vspace{2mm}
    \centering
    \includegraphics[width=\linewidth, trim={50 30 90 125}, clip]{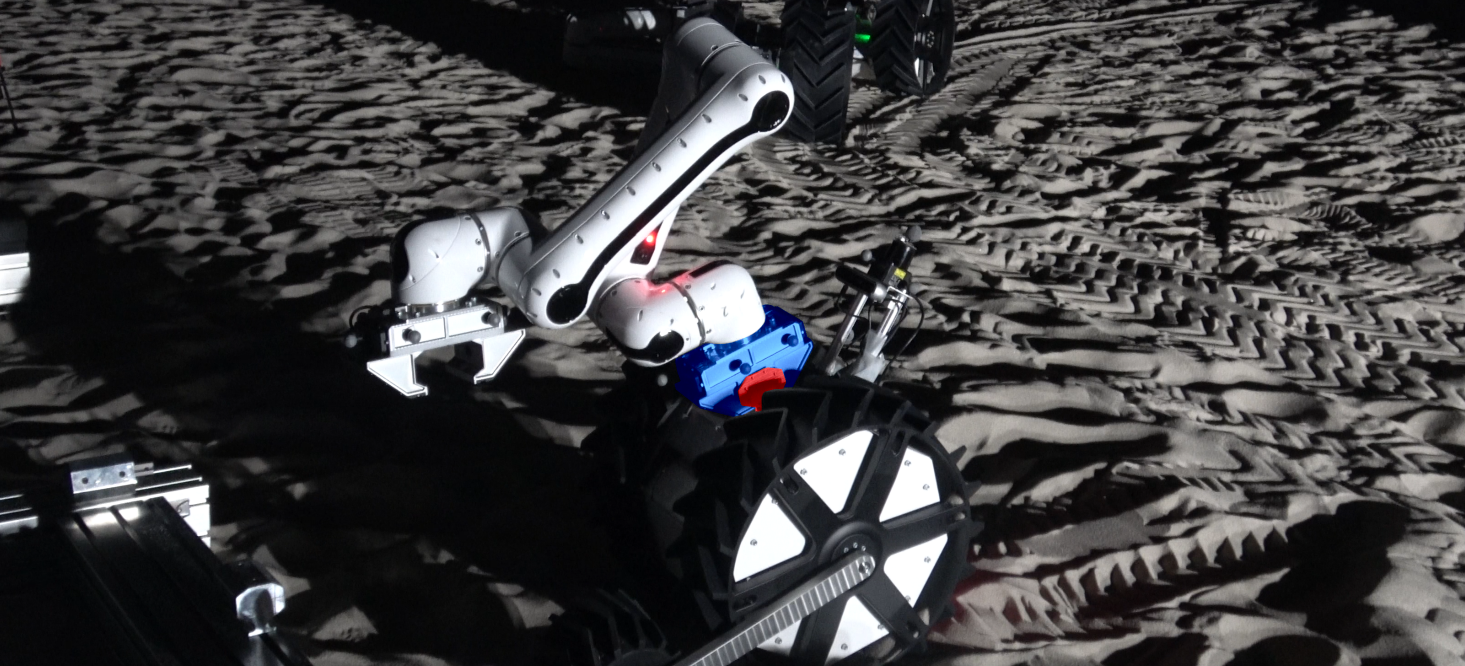}
    \caption{Reconfiguration}
    \label{fig:exp_reconfig}
  \end{subfigure}
    \caption{Test stages of the \MB Minimal self-assembly. Blue masking highlights the end-effector (Limb V2),
    Red masking highlights the target (grasping point of Wheel V2).}
  \label{fig:exp_stages}
\end{figure}

\begin{figure}[b]
  \centering
  \begin{subfigure}[b]{0.49\columnwidth}
    \centering
    \includegraphics[width=\linewidth, trim=110 55 280 60, clip]{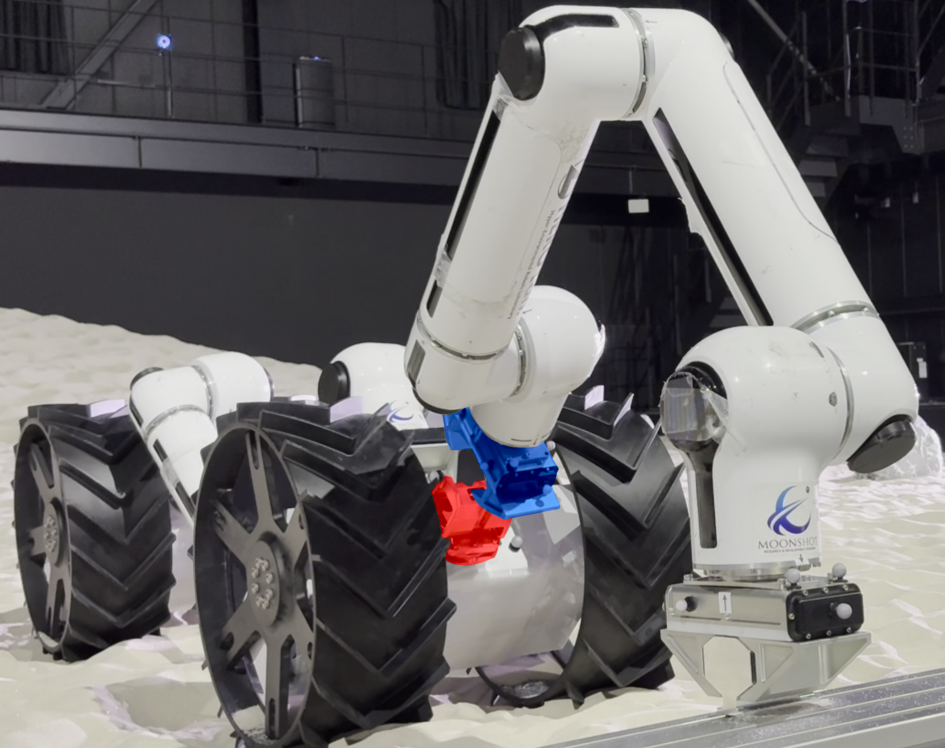}
    \caption{Alignment}
    \label{fig:drag_align}
  \end{subfigure}
  \begin{subfigure}[b]{0.49\columnwidth}
    \centering
    \includegraphics[width=\linewidth, trim=110 55 280 60, clip]{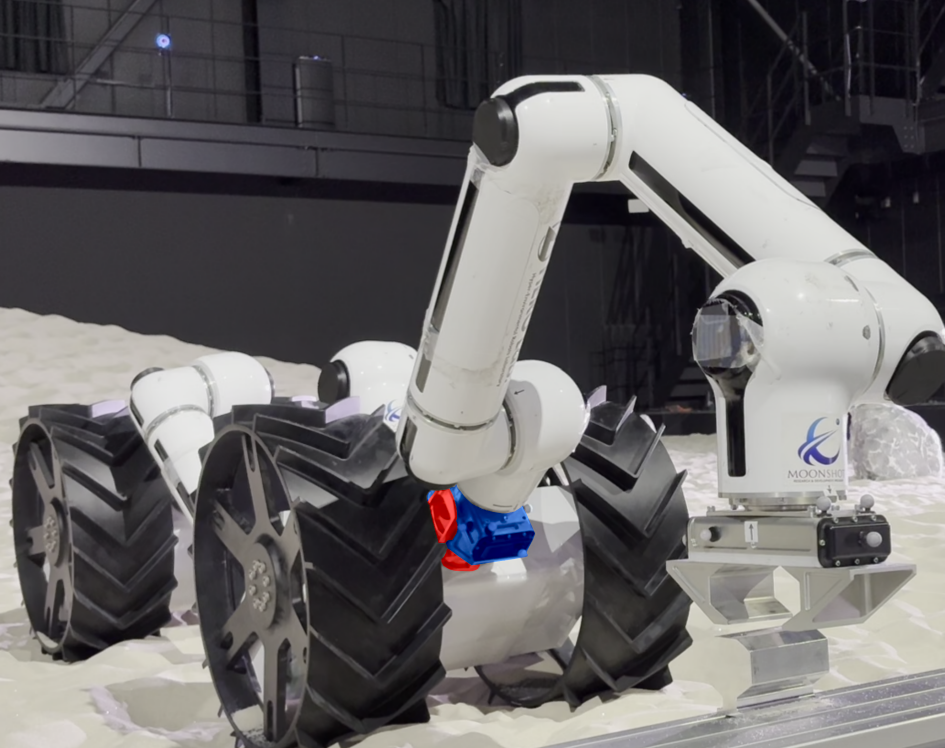}
    \caption{Grasping and Lifting Up}
    \label{fig:drag_grasp}
  \end{subfigure}
  \caption{\MB{} Limb V1 executing the self-assembly sequence toward the \MB Wheel V1
    target to assemble the Dragon configuration. End-effector is marked in blue and target in red.}
  \label{fig:limb1_dragon}
\end{figure}

All experiments were self‐assembly trials of both the \MB{} {\it Minimal} (\cref{fig:exp_reconfig})
and \MB{} {\it Dragon} (\cref{fig:drag_grasp}) configurations of the \MBs' robotic system,
conducted in the Space Exploration Field at JAXA Sagamihara campus.
A sand‐filled testbed simulated regolith conditions, and a pallet fixture anchored each \MB{} limb
in place (\cref{fig:exp_stages,fig:limb1_dragon}).
\cref{fig:exp_stages} shows \MB Limb 2 self‐assembling to the \MB{} Minimal configuration using
the \MB Wheel V2 target, while \cref{fig:limb1_dragon} illustrates \MB Limb V1 assembling to the \MB Dragon
configuration against the \MB Wheel V1 target. These trials demonstrate the controller’s
agnosticism to both manipulator and target geometry.

We used an OptiTrack motion-capture system (16 cameras: PrimeX 41/22) to track
both the end-effector and the wheel target at 180 Hz, achieving sub-millimeter
accuracy after a standard wand calibration \cite{optitrack}. For comparability,
the wheel target remained fixed across runs and both limbs were mounted with
identical geometry.

An operator supervised all trials for safety \cite{mishra_enhancing_2025, leeper_strategies_2012}:
they set the initial “safe” pose, triggered stage transitions (align/grasp), and
could pause or abort; motion within stages was autonomous from external pose feedback.

The controller continuously processed live pose measurements to generate motion commands
automatically, without robot specific information.
For each limb and each controller version, we performed three independent trials (12 runs total)
under identical setup conditions to ensure robust performance comparison.

\subsection{Task Sequence and Metrics}

We structured each trial into four sequential stages to mimic a real self‐assembly operation:

\begin{enumerate}
  \item \textbf{Initial safe pose:}
    The manipulator is first moved by the operator into a predefined “safe” configuration,
    which serves as the starting posture for all subsequent autonomous motions (see \cref{fig:limb2_wheel}).
  \item \textbf{Alignment:}
    For collision avoidance and to accommodate varying target geometries, the controller drives the end-effector
    to a pose offset 150 mm above the true grasp point. This offset approach ensures a consistent, safe approach
    regardless of wheel shape (see \cref{fig:exp_align,fig:drag_align}).
  \item \textbf{Grasping:}
    From the offset position, the controller is re-invoked without the vertical offset, allowing the end-effector
    to descend precisely to the actual grasping point. Once in position, the gripper closes to secure the module
    to the limb (see \cref{fig:exp_grasp,fig:drag_grasp}).
  \item \textbf{Reconfiguration:}
    After grasping, the opposite limb or mounting fixture releases its gripper, the newly attached end-effector
    lifts the module, and the arm executes a joint-space transition into the target assembly posture
    (Minimal or Dragon), completing the self-assembly sequence, as shown in \cref{fig:exp_reconfig,fig:drag_grasp}.
\end{enumerate}

For each stage we recorded:
\begin{itemize}
    \item \textbf{Full trajectory log:} Complete end‐effector pose history \((x_e(t),q_e(t))\) sampled
        at 10 Hz, allowing detailed offline analysis of path length, curvature, and oscillations.
  \item \textbf{Completion time:} Measured from command issuance to convergence within error bounds.
  \item \textbf{Final alignment error:} The residual translational \(\Delta d\) and rotational
    \(\Delta\theta\) at the end of the alignment stage.
\item \textbf{Hypersphere parameters:} Logged both the base radii \(\Delta_t,\Delta_r\) \cref{eq:base_radii} and the
    effective radii \(\Delta_t^{\rm eff},\Delta_r^{\rm eff}\) \cref{eq:eff_radii} at each high‐level update.
\end{itemize}

\subsection{Results and Comparison}
Table~\ref{tab:alignment_summary} summarizes the overall alignment performance
for both limbs and controller versions. Each entry reports the total alignment
duration, the final translational error \(\Delta d\), and the final rotational
error \(\Delta\theta\), averaged over three trials.

\begin{table}[b]
  \caption{Alignment performance (mean ± s.d., \(n=3\) per condition).}
  \label{tab:alignment_summary}
  \centering
  \renewcommand{\arraystretch}{1.2}
  \begin{tabular}{l c c c}
    \hline
    \textbf{Version / Limb} & \textbf{Duration (s)} & \(\mathbf{\Delta d}\) \textbf{(mm)} & \(\mathbf{\Delta\theta}\) \textbf{(deg)} \\
    \hline
    Version 1 (Limb V1) & 92.90 ± 1.98 & 5.93 ± 0.14 & 0.35 ± 0.04 \\
    Version 2 (Limb V1) & 57.86 ± 8.23 & 4.51 ± 0.64 & 0.59 ± 0.07 \\
    Version 1 (Limb V2) & 52.00 ± 4.18 & 8.78 ± 0.83 & 0.34 ± 0.31 \\
    Version 2 (Limb V2) & 46.83 ± 2.63 & 4.57 ± 0.77 & 1.49 ± 0.62 \\
    \hline
  \end{tabular}
\end{table}

\begin{figure}[t]
\vspace{2mm}
  \centering
  \begin{subfigure}[b]{0.49\columnwidth}
    \includegraphics[width=\linewidth,trim=30 14 0 38, clip]{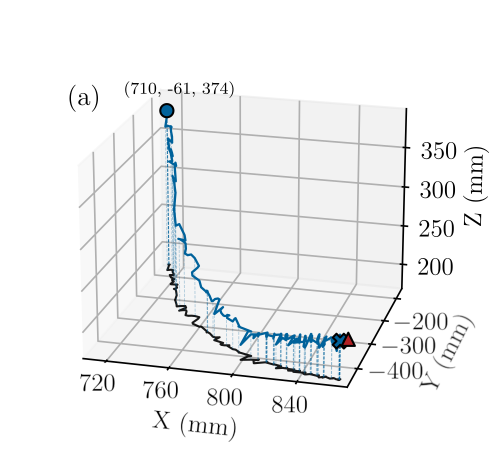}
    \caption{Limb V1 + Controller Ver. 1}
    \label{fig:traj_a}
  \end{subfigure}\hfill
  \begin{subfigure}[b]{0.49\columnwidth}
    \includegraphics[width=\linewidth,trim=30 14 0 38, clip]{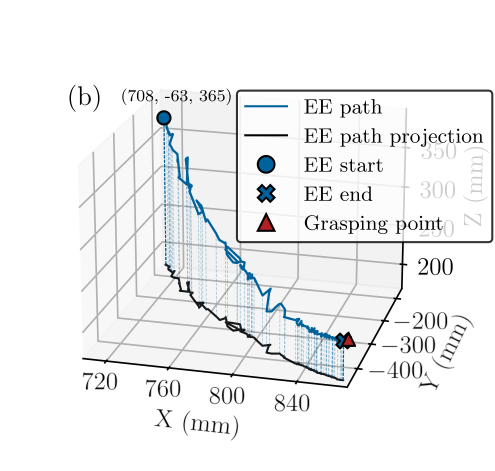}
    \caption{Limb V1 + Controller Ver. 2}
    \label{fig:traj_b}
  \end{subfigure}

    \vspace{2mm}
  % (c) Limb 2, Version 1
  \begin{subfigure}[b]{0.49\columnwidth}
    \includegraphics[width=\linewidth,trim=30 14 0 38, clip]{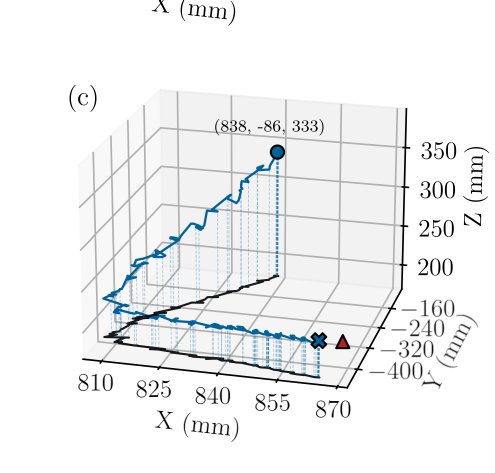}
    \caption{Limb V2 + Controller Ver. 1}
    \label{fig:traj_c}
  \end{subfigure}\hfill
  % (d) Limb 2, Version 2
  \begin{subfigure}[b]{0.49\columnwidth}
    \includegraphics[width=\linewidth,trim=30 14 0 38, clip]{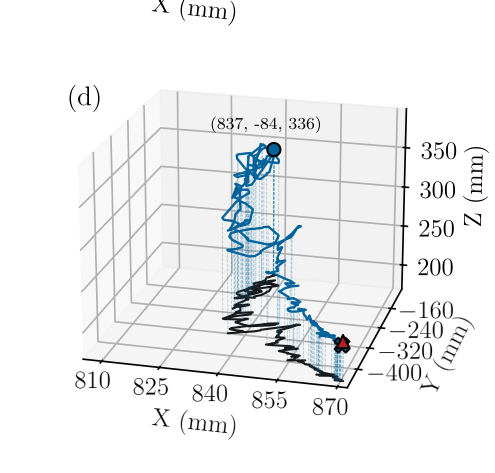}
    \caption{Limb V2 + Controller Ver. 2}
    \label{fig:traj_d}
  \end{subfigure}

 \caption{Alignment-only trajectories for both \MB limbs converging on a stationary wheel at \((871,\,-459,\,221)\) mm using both controller versions. 
    All trials used the same 30\,Hz high-level loop.}
  \label{fig:trajectories}
\end{figure}

\subsection{Trajectory Analysis and Controller Validation}
\label{sec:analysis}

\paragraph{Version 1 (two‐stage LERP) on both limbs}
Figures~\ref{fig:traj_a} and \ref{fig:traj_c} show that the stepwise LERP strategy
yields very smooth, monotonic convergence with negligible lateral wobble.
Although Limb V2 follows a slightly longer arc due to its kinematics, its paths remain
tightly clustered around the mean, demonstrating reliable performance across different geometries.

\paragraph{Version 2 (adaptive hypersphere clamp)}
On Limb V1 (\cref{fig:traj_b}), the continuous‐velocity method produces a more direct path,
reducing execution time while maintaining stability -- only minor oscillations occur near the end.
On Limb V2 (\cref{fig:traj_d}), higher sensor jitter and mechanical play lead to more noticeable wobble,
but the overall trajectory remains shorter than with Version 1, illustrating the clamp’s ability to adapt
to less ideal conditions without compromising convergence.

\paragraph{Summary and Trade-Offs}
Controller Version 2 cuts alignment time by roughly one quarter on average (Tab.~\ref{tab:alignment_summary})
and lowers positional residuals, at the expense of a small increase in rotational error.
Version 1, conversely, offers slower but exceptionally stable motions with near-zero drift.
These complementary behaviors allow mission planners to choose between maximum speed or maximum precision.

\begin{figure}[t]
\vspace{2mm}
  \centering
  \begin{subfigure}[b]{\columnwidth}
    \centering
    \includegraphics[width=\linewidth, trim=10 597 225 39, clip]{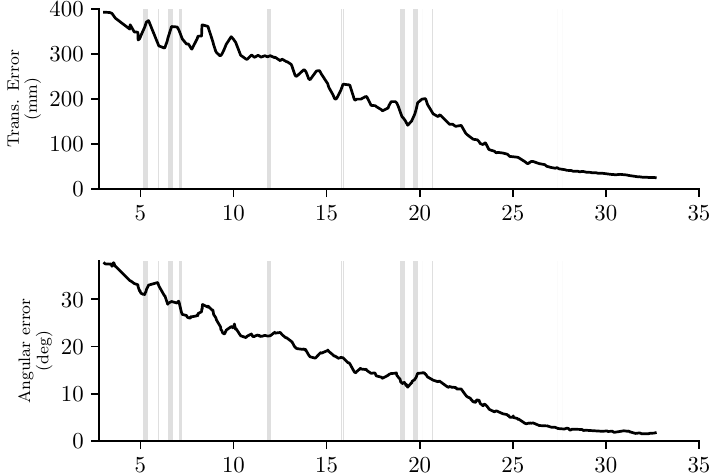}
    \caption{Translational error \(\Delta d(t)\) and rotational error \(\Delta\theta(t)\) during alignment.}
    \label{fig:error_velocity_errors}
  \end{subfigure}
  \begin{subfigure}[b]{\columnwidth}
      \vspace{2mm}
    \centering
    \includegraphics[width=\linewidth, trim=10 442 225 279, clip]{./fig/alignment_dynamics.pdf}
      \caption{Original Raw vs.\ Clamped Velocity magnitudes.}
    \label{fig:error_velocity_vels}
  \end{subfigure}

  \caption{Time‐series of alignment dynamics for \MB{} Limb V2 under Controller Version 2.}
  \label{fig:error_velocity}
\end{figure}

\paragraph{Velocity Clamping Dynamics}
\cref{fig:error_velocity_errors} confirms that translational and rotational errors converge in unison,
thanks to the simultaneous clamping in \cref{eq:ellipsoid_norm}.
\cref{fig:error_velocity_vels} illustrates how raw commands exceeding the dynamic bounds from
\cref{eq:clamped_vel} are scaled back into the hypersphere, preventing either axis from dominating.
Zero‐hold intervals triggered by stability checks appear as flat segments in the clamped velocities,
and the filter in \cref{eq:smooth} then gently ramps commands back up, ensuring smooth, safe recovery.

%%%%%%%%%%%%%%%%%%%%%%%%%%%%%%%%%%%%%%%%%%%%%%%%%%%%%%%%%%%%%%%%%%%%%%%%%%%%%%%%
%%%%%%%%%%%%%%%%%%%%%%%%%%%%%
\section{Discussion}
\label{sec:discussion}

\subsection{System Agnosticism}
A key strength of our framework is its hardware- and task-agnostic design.
Both Controller Version 1 (Sec.~\ref{sec:version1}) and Version 2 (Sec.~\ref{sec:version2})
were applied without modification to two distinct 7-DOF \MB limbs (V1 and V2)
and two different assembly targets (Minimal and Dragon).
This reuse demonstrates seamless portability across varying kinematics and docking geometries,
significantly reducing integration effort when deploying to new modular configurations.
By relying solely on generic end-effector and target pose feedback, our alignment logic
can be paired with any motion-planning or middleware stack, accelerating research and
collaboration in both modular space robotics and conventional robotic applications alike.
With appropriate tuning of thresholds and shrink factors, the same controller can support
tasks ranging from industrial pick-and-place to delicate surgical manipulation.

\subsection{Controller Selection Trade-Offs}
Our experiments (Sec.~\ref{sec:analysis}) reveal a clear trade-off between the two alignment strategies.
Version 1’s step-and-settle LERP approach yields ultra-stable, near-drift-free motions
but incurs longer alignment times (up to 90 s), making it well suited for tasks where
maximum smoothness and predictability are paramount.
Version 2’s continuous adaptive clamp reduces alignment time by roughly 20-25 \% (Tab.~\ref{tab:alignment_summary})
while keeping final position errors in the few-millimeter range, at the expense of modest oscillations on noisier limbs.
This variant excels when speed is critical, or when operating on platforms with lower intrinsic noise.

By offering both modes within the same software layer, mission planners can tailor the controller
to specific operational priorities -- whether prioritizing rapid convergence in a time-sensitive scenario,
or guaranteeing minimal lateral deviation in delicate assembly maneuvers.
%%%%%%%%%%%%%%%%%%%%%%%%%%%%%%%%%%%%%%%%%%%%%%%%%%%
\section{Conclusion}

We introduced a robot-agnostic end-effector alignment controller that closes the loop on
external pose feedback and regulates motion with an adaptive hypersphere clamp. Two variants,
a discrete step-and-settle strategy (\cref{sec:version1}) and a continuous velocity-based
strategy (\cref{sec:version2}), were validated on two \MB limbs and two assembly targets in field
trials (\cref{sec:experiment}), revealing task-dependent trade-offs (\cref{sec:analysis}).

On our MoonBot hardware, non-idealities were substantial: some joints exhibited up to
\(10^{\circ}\) mechanical backlash, and accumulated flex and wobble produced end-effector
position errors on the order of \(10\,\text{cm}\). These effects varied across robots and over
time. By leveraging exteroceptive pose measurements and adaptively clamping translational and
rotational velocities, the controller maintained stable, smooth convergence under time-varying
uncertainties. In practice, the stepwise variant favors maximal smoothness and predictability,
whereas the continuous variant provides quicker convergence on noisier systems. Because both share
a simple pose-only interface, the method deploys quickly across platforms, including terrestrial
applications that require robust closed-loop pose correction.

Looking ahead, we plan to enrich the stability model with additional sensing modalities such as force/torque data
and stereo vision, allowing contact-aware and obstacle-sensitive behaviors. We will explore online learning techniques
to automatically adjust shrink-factor weights based on performance metrics, further reducing manual parameter tuning.
Finally, we aim to incorporate this alignment controller as a component within the Motion Stack
open-source ecosystem. In doing so, we envision enabling not only self-assembly and reconfiguration,
but also large-scale autonomous construction, resource transport, and maintenance operations on the
lunar surface and beyond.

\bibliographystyle{IEEEtran}
\bibliography{bib}

\end{document}